\documentclass[11pt]{article}

\usepackage[preprint]{acl}

\usepackage{times}
\usepackage{latexsym}
\usepackage{amsmath}
\usepackage{amssymb}
\usepackage{algorithm}
\usepackage{algpseudocode}

\usepackage{multirow}

\usepackage[T1]{fontenc}
\usepackage[utf8]{inputenc}

\usepackage{microtype}

\usepackage{inconsolata}

\usepackage{graphicx}
\usepackage{booktabs}
\usepackage[table]{xcolor}
\usepackage{array}
\usepackage{tabularx}

\title{Analytic Concept-Centric Memory for Agentic Embodied Manipulation}

\author{
  Mingyang Sun$^{1,2,4}$
  \quad
  Xiujian Liang$^{2,3}$
  \quad
  Jiude Wei$^{5}$
  \quad
  Qichen He$^{2,5}$
  \\
  \textbf{Donglin Wang}$^{4}$
  \quad
  \textbf{Cewu Lu}$^{2,5}$
  \quad
  \textbf{Jianhua Sun}$^{5}$\thanks{Corresponding author.}
  \\[0.5em]
  $^{1}$Zhejiang University, Hangzhou, China \quad
  $^{2}$Shanghai Innovation Institute, Shanghai, China \\
  $^{3}$Fudan University, Shanghai, China \quad
  $^{4}$Westlake University, Hangzhou, China \\
  $^{5}$Shanghai Jiao Tong University, Shanghai, China
  \\[0.2em]
  \small{\href{mailto:sunmingyang@zju.edu.cn}{\texttt{sunmingyang@zju.edu.cn}},\quad \href{mailto:gothic@sjtu.edu.cn}{\texttt{gothic@sjtu.edu.cn}}}
}

\begin{document}
\maketitle
\begin{abstract}
Long-horizon embodied manipulation requires agents to remember persistent objects, track changing scene states, and reuse prior interaction knowledge. However, existing agent memories are often stored as unstructured histories or embedding-based records, making it difficult to retrieve manipulation-relevant object parts, physical states, action effects, and executable skills. We propose an \textbf{analytic concept-centric memory} framework for agentic embodied manipulation. Our memory organizes experience around structured analytic concepts, where objects are represented by semantic parts, parametric templates, grounded poses, affordances, and manipulation states. It further connects object and scene memories with transition memory for action-induced state changes and skill memory for template-grounded and policy-grounded execution. At runtime, the agent performs structured coarse-to-fine retrieval to identify relevant objects, states, transitions, and skills, supporting state-consistent reasoning and skill reuse. Experiments on memory-dependent manipulation, articulated-object generalization, real-world memory evaluation, and ablations show that our approach improves task completion, retrieval accuracy, object re-identification, and cross-object skill generalization over unstructured and embedding-based memory baselines.
\end{abstract}

\section{Introduction}

Embodied agents are expected to perform long-horizon robotic manipulation tasks in dynamic physical environments~\citep{fang2019scene, gervet2023navigating, fung2025embodied}. Such tasks require agents not only to perceive the current scene and execute actions, but also to remember persistent objects, track changing states, reuse previous interaction knowledge, and recover from execution failures. Without an explicit memory mechanism, an agent may repeatedly rediscover the same object, lose track of state changes, or fail to exploit prior experience when facing structurally similar objects and tasks~\citep{hu2025memory}.

Recent agent memory systems have explored storing interaction histories, scene observations, textual summaries, visual embeddings, or episodic traces. While these memories provide useful context, they are often insufficient for robotic manipulation. Unstructured memories make it difficult to retrieve task-relevant object and state information efficiently. Embedding-based memories~\citep{zhang2025mem2ego, wang2025m+} may capture similarity, but often lack explicit access to object parts, affordances, and action effects. Scene-level ~\citep{wang2025karma} or episode-level memories~\citep{zhu2024retrieval} usually describe what happened, but do not directly represent how object structures support manipulation or how actions change object states. 
% As a result, existing memory systems provide limited support for object-level persistence, part-level reasoning, and compositional generalization in embodied tasks.

Current approaches to long-horizon embodied tasks often follow two complementary routes: end-to-end VLA policies~\citep{memoryvla,li2026remem} provide strong visuomotor execution but keep task reasoning and memory largely implicit, while agentic systems~\citep{anwar2025remembr,lei2026towards,tan2025roboos} make planning and tool use explicit but often depend on manually specified skills or task-specific robot APIs. However, neither line provides an explicit memory interface that is simultaneously
language-addressable, object-structured, and execution-grounded. As a result, the agent
may retrieve semantically relevant memories that are physically inapplicable, or select
skills without verifying object states and part-level constraints.

To address these limitations, we propose an \textbf{Analytic Concept-centric Memory}~(\textbf{ACM}) for embodied agents. Our key motivation is that manipulation-relevant memory should be organized around structured object concepts rather than raw observations or flat experience records. An analytic concept represents an object through semantic parts, parametric templates, geometric attributes, and instance-specific grounding~\citep{NEURIPS2024_89d19544, Sun_2025_ICCV}. This structured representation allows the agent to store what an object is, how it is composed, where it appears in a scene, how its state changes under actions, and which executable skills can be applied to it. By making these elements explicit, analytic concept memory provides a bridge between perception, memory retrieval, task reasoning, and robot execution.

% Our memory system consists of both stable knowledge components and dynamically updated experience components. The \textbf{semantic library} stores category-level priors such as object parts, affordances, and functional attributes. The \textbf{template library} stores reusable parametric structures for representing common object parts and shapes. The \textbf{object memory} maintains grounded object instances with persistent identities and structured part-level descriptions. The \textbf{scene memory} records object poses, states, visibility, and spatial relations over time. The \textbf{transition memory} stores action-induced state changes by linking object parts, actions, preconditions, and effects. The \textbf{procedural memory} stores executable skills, motion primitives, or policy references, allowing the agent to map high-level decisions to physical actions.
Our memory system combines stable template-level knowledge with dynamic experience memory.
Stable libraries provide template synopses and parametric analytic templates, while dynamic
memories maintain grounded objects, scene states, action transitions, and executable skills.
% At runtime, the agent retrieves these entries for state-consistent reasoning and updates
% them with execution outcomes.
At runtime, the agent queries memory conditioned on observations and instructions to ground the current scene, predict action effects, and select or parameterize executable skills. Execution outcomes are then written back to update memory, maintaining state consistency and accumulating reusable experience over episodes.

Our contributions are threefold:
\begin{itemize}
    \item We propose a memory-augmented embodied agent pipeline for long-horizon manipulation, where task execution is grounded on persistent memory around structured object concepts rather than raw observations or unstructured experience records.
    
    \item We introduce an analytic concept-centric memory organization that links object-centric analytic concepts, action transitions, and executable skills, enabling persistent object reasoning and compositional manipulation.
    
    \item We design a unified skill memory that bridges object-centric manipulation priors and task-centric learned policy skills, allowing the agent to connect structured memory retrieval with executable robot behaviors.
\end{itemize}

We evaluate our approach on memory-dependent manipulation tasks that require object persistence, state tracking, action-effect reasoning, and generalization across structurally similar objects. Analytic concept memory improves memory retrieval, object re-identification, and task success rate.

\section{Related Work}

\subsection{Embodied Agentic Systems}

Recent embodied AI systems increasingly adopt agentic frameworks for long-horizon and language-conditioned manipulation tasks~\citep{salimpour2025towards,liang2025large}. Hierarchical embodied agents typically decompose high-level goals into executable subtasks or reusable skills, combining planning modules with low-level robot controllers. Early systems such as SayCan~\citep{ahn2022can} and Code as Policies~\citep{liang2023code} leverage language models to select skills, invoke APIs, or synthesize executable programs for robot control. More recent approaches integrate stronger visual-language-action (VLA)~\citep{brohan2022rt,kim2024openvla} policies into agentic execution loops, enabling improved planning, replanning, and skill composition for long-horizon manipulation~\citep{yang2025agenticrobotbraininspiredframework,lei2026towards}.
Recent works further explore modular policy composition and skill-centric execution.  ThinkAct~\citep{huang2026thinkact} combines reasoning modules with low-level executors for closed-loop embodied execution, while LiLo-VLA~\citep{yang2026lilo} demonstrates that modular object-centric policies can improve compositional manipulation through dynamic replanning and skill reuse.

These works highlight the importance of hierarchical reasoning and modular execution in embodied agents. However, existing systems primarily focus on planning and policy orchestration, while memory is often limited to short interaction histories, latent context windows, or unstructured retrieval mechanisms. 
% As a result, agents still lack persistent object-centric memory for tracking object states, representing manipulation dynamics, and reusing structured interaction knowledge across long-horizon tasks. Our work addresses this limitation by organizing embodied memory around analytic concepts, enabling structured retrieval, transition reasoning, and executable skill grounding.
\subsection{Memory for Agents}

Recent advances in agent memory have established memory as a fundamental component of LLM-based agents, enabling long-horizon reasoning, continual adaptation, and environment interaction. Existing studies mainly categorize memory into token-level, parametric, and latent forms, further supporting factual, experiential, and working memory functionalities~\citep{hu2025memory}. Most current systems adopt token-level memory, where dialogue histories, trajectories, summaries, or experience buffers are explicitly stored and retrieved, as exemplified by frameworks~\citep{packer2023memgpt,chhikara2025mem0,zhang2025darwin}.

However, most existing memory systems target language/web agents or simplified simulations, and do not directly meet the needs of embodied robotics~\citep{zheng2025skillweaver, ye2025agentfold, fang2025memp}. In real-world interaction, naively storing multimodal trajectories incurs high storage and retrieval overhead under real-time constraints. Moreover, many representations remain semantic-centric and weakly connected to grounded affordances and executable manipulation skills.

As a result, existing approaches often memorize task-specific experiences but struggle to generalize manipulation knowledge across diverse objects and environments. These limitations motivate our analytic concept-centric memory, which aims to unify object semantics, environment understanding, and manipulation skills into a compact and transferable memory representation for embodied agents.

\section{Preliminaries}

\paragraph{Language-conditioned embodied manipulation.}
We study long-horizon language-conditioned embodied manipulation, where an agent receives an instruction $L$ and a sequence of visual observations $\{o_t\}$. We model execution hierarchically. At the task level, $L$ is decomposed into subtasks $\mathbf{l}=\{l_1,\ldots,l_N\}$; at the execution level, each subtask $l_i$ is realized by a sequence of low-level robot actions $\boldsymbol{a}_i=\{a_{i,1},\ldots,a_{i,K_i}\}$. The full trajectory is $\boldsymbol{a}=\{\boldsymbol{a}_1,\ldots,\boldsymbol{a}_N\}$, where each $a_{i,j}$ denotes an executable robot command, such as an end-effector pose, trajectory command, or controller input.

In our setting, subtask execution is conditioned on an external memory $\mathcal{M}$:
\[
\boldsymbol{a}_i = \pi(L, o_t, l_i, \mathcal{M}),
\]
where $\mathcal{M}$ provides persistent object identities, part-level structure, evolving scene state, action-effect knowledge, and executable skills beyond the current observation.

\paragraph{Analytic concept representation.}
An analytic concept~\citep{NEURIPS2024_89d19544, Sun_2025_ICCV} is a structured, manipulation-aware representation of a physical object. Instead of representing an object as a raw visual instance or a single embedding, it decomposes the object into semantic parts, geometric parameters, poses, affordances, and manipulation templates.

Formally, an object $O$ is represented as
\[
O = \mathcal{O}(c, \{P_i\}_{i=1}^{M}),
\]
where $c$ denotes the object category and $P_i$ denotes the $i$-th semantic part. Each
part is associated with a template type $\tau_i$ and instantiated as
\[
P_i = \mathcal{P}(\tau_i, \Theta_i, T_i, A_i, \mathcal{F}_{\tau_i}),
\]
where $\tau_i$ links the part to a reusable template in the template library, $\Theta_i$
denotes the estimated geometric parameters, $T_i$ denotes the grounded 6D pose, $A_i$
denotes part-level affordances, and $\mathcal{F}_{\tau_i}$ denotes the manipulation
templates supported by this template type.

This representation makes object structure explicit while keeping the symbols reusable:
template types define reusable part and manipulation structures, while grounded
parameters and poses instantiate them into scene-specific object concepts.

% \section{Method}

\begin{figure*}[t]
    \centering
    \includegraphics[width=\linewidth]{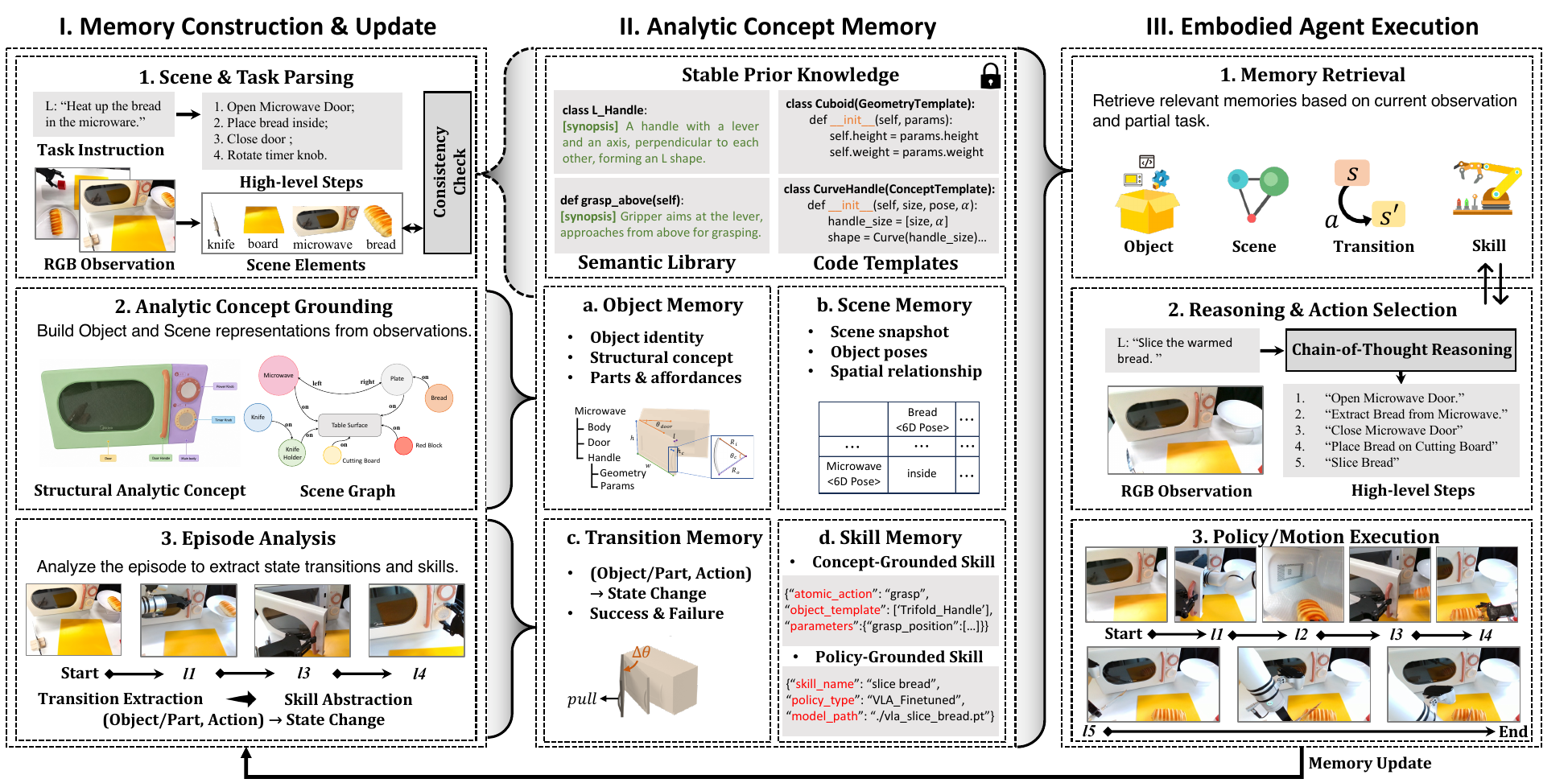}
    \caption{Overview of the proposed analytic concept-centric memory system. Stable semantic and template libraries provide reusable priors, while dynamic object, scene, transition, and skill memories store grounded instances, evolving states, and action effects. At runtime, the agent retrieves relevant entries to ground subtasks and select/parameterize executable skills, and then writes back observations and outcomes to continuously update memory.}
    \label{fig:overview}
\end{figure*}

\section{Method}

We propose an analytic concept-centric memory framework for long-horizon embodied manipulation.
As shown in Figure~\ref{fig:overview}, the framework maintains a structured memory
$\mathcal{M}$ that connects object-centric concepts, scene states, action transitions, and
executable skills. At a high level, the agent first grounds observations into analytic
concepts, retrieves task-relevant memory entries for the current subtask, selects or
parameterizes an executable skill, and then writes execution outcomes back to memory.
This closed loop allows the agent to maintain persistent object identities, reason about
state changes, and reuse manipulation knowledge across related objects and tasks.

\subsection{Analytic Concept Memory}

The memory $\mathcal{M}$ is organized around analytic concepts and consists of stable prior knowledge
and dynamic experience memory :
\[
\{\mathcal{M}^{sem}, \mathcal{M}^{temp},
\mathcal{M}^{obj}, \mathcal{M}^{scene},
\mathcal{M}^{trans}, \mathcal{M}^{skill}\}.
\]
The stable components, $\mathcal{M}^{sem}$ and $\mathcal{M}^{temp}$, provide
language-level template descriptions and parametric analytic templates, respectively. The dynamic components store grounded object
instances, evolving scene states, action-induced transitions, and executable skills
collected during interaction.

\subsubsection{Stable Prior Knowledge}

% Stable prior knowledge provides reusable descriptions and templates that are shared
% across tasks and episodes. It contains two complementary libraries. The semantic library
% describes the meaning and intended use of templates in natural language, while the
% template library stores the corresponding parametric structures and manipulation
% functions. In this work, both libraries are initialized before execution and kept fixed
% during evaluation; online observations and execution outcomes are written to dynamic
% memory.
Stable prior knowledge contains two fixed libraries shared across episodes:
a semantic library of template synopses and a template library of parametric analytic
templates. Online observations and outcomes are written only to dynamic memory.

\paragraph{Semantic Library.}
The semantic library $\mathcal{M}^{sem}$ stores natural-language synopses for templates
in $\mathcal{M}^{temp}$. Each entry is indexed by the same template type $\tau$ as the
template library:
\[
m^{sem}_{\tau} = (\tau, d_{\tau}, a_{\tau}, u_{\tau}),
\]
where $d_{\tau}$ is a concise textual description of the template, $a_{\tau}$ describes
its interaction affordances, and $u_{\tau}$ describes its intended functional use.
For example, the synopsis of an \texttt{L\_Handle} template describes it as a handle
with a lever and an axis, where the lever supports grasping and the axis indicates the
direction for pulling or rotation. These synopses provide a language-level interface for
selecting template types during concept grounding, memory retrieval, and skill selection.

\paragraph{Template Library.}
The template library $\mathcal{M}^{temp}$ stores reusable analytic templates indexed by
template type $\tau$:
\[
m^{temp}_{\tau}=(\tau,\Omega_{\tau},\mathcal{F}_{\tau}),
\]
where $\Omega_{\tau}$ denotes the valid parameter space of the template, and
$\mathcal{F}_{\tau}$ is the set of supported manipulation functions. During grounding,
scene parsing modules, including object detection, part segmentation, pose estimation,
and parameter estimation, bind a detected part to a template type $\tau_i$ and estimate
scene-specific values $(\hat{\Theta}_i,\hat{T}_i)$, producing a grounded part concept
$P_i$. The manipulation
functions in $\mathcal{F}_{\tau}$ are implemented using the atomic primitives defined in
Appendix~\ref{sec:appendix_primitives}. Thus, templates provide reusable symbolic and
geometric blueprints, while object memory stores their scene-specific executable
instantiations.
% The semantic and template libraries are linked by the template type $\tau$: the semantic
% library explains what a template means and when it should be used, while the template
% library specifies how the template is instantiated and executed. This separation allows
% language-based reasoning modules to query human-readable template descriptions, while
% the execution modules operate on structured parameters and manipulation functions.

\subsubsection{Dynamic Experience Memory}

Dynamic memory records grounded information accumulated during interaction. Unlike stable
prior knowledge, dynamic memory is updated online as the agent observes the scene,
executes actions, and verifies their outcomes.

\paragraph{Object Memory.}
Object memory $\mathcal{M}^{obj}$ maintains persistent object-level analytic concepts
grounded in the current environment:
\[
m^{obj}_k=(id_k,c_k,\{P_{k,i}\}_{i=1}^{M_k},T_k,s_k,h_k).
\]
Here, each part $P_{k,i}$ reuses a template type $\tau_{k,i}$ from
$\mathcal{M}^{temp}$ and binds it with scene-specific parameters, poses, affordances, and
supported manipulation functions estimated during grounding. Thus, each object memory
entry is an executable AC instance: it preserves a persistent identity while retaining
the template-level structure needed for retrieval and skill execution.

\paragraph{Scene Memory.}
Scene memory $\mathcal{M}^{scene}$ tracks the evolving environment state. We represent a
scene snapshot as a graph
\[
G_t = (V_t, E_t),
\]
where nodes $V_t$ correspond to object instances and edges $E_t$ encode spatial,
visibility, containment, support, and state relations. Scene memory enables multi-step planning by retaining relations not fully observable at the current timestep.

\paragraph{Transition Memory.}
Transition memory $\mathcal{M}^{trans}$ stores action-induced state changes over
grounded objects and parts:
\[
m^{trans}=(id_k, P_{k,i}, u, s_{\mathrm{pre}}, s_{\mathrm{post}}, r),
\]
where $id_k$ is the persistent target object, $P_{k,i}$ is the grounded target part, $u$
is the executed skill or primitive action, $s_{\mathrm{pre}}$ and $s_{\mathrm{post}}$
are pre- and post-action states, and $r$ is the execution outcome. This links action
effects directly to object-memory AC instances for later precondition checking and
effect prediction.

\paragraph{Skill Memory.}
Skill memory $\mathcal{M}^{skill}$ stores executable procedures:
\[
m^{skill}_u=(u,type,\tau,target,pre,\Phi,\pi_{\mathrm{exec}},\eta),
\]
where $u$ identifies the skill, $type$ denotes template- or policy-grounded execution,
$\tau$ is the associated template type when applicable, $target$ specifies the required
object/part schema, $pre$ denotes applicability conditions, $\Phi$ contains bound
parameters, $\pi_{\mathrm{exec}}$ is the execution backend, and $\eta$ stores execution
statistics. Template-grounded skills bind $f\in\mathcal{F}_{\tau}$ to a grounded part
$P_{k,i}$, while policy-grounded skills invoke learned policies for behaviors that are
hard to express with explicit templates.

\subsection{Memory Construction \& Update}

This module converts instructions, observations, and execution outcomes into memory
updates. Given $(L,o_t,\mathcal{M}_t)$, it parses the task, grounds observed entities,
resolves object identities, and updates object, scene, transition, and skill memories.

\subsubsection{Task and Scene Parsing}

The first stage extracts a task-scene context from the instruction and the current
observation. The instruction $L$ is decomposed into high-level subtask candidates
$\mathbf{l}=\{l_1,\ldots,l_N\}$. Meanwhile, the observation $o_t$ is parsed to detect
task-relevant objects, semantic parts, and environmental elements. The parser also
extracts preliminary spatial, state, and functional relations, such as
\texttt{on}, \texttt{inside}, \texttt{open}, or \texttt{graspable}. The output is a
structured context
\[
\mathcal{Q}_t = (\mathbf{l}, \mathcal{O}_t, \mathcal{R}_t),
\]
where $\mathcal{O}_t$ denotes the set of observed entities and $\mathcal{R}_t$ denotes
their preliminary relations. 
\subsubsection{Object and Scene Concept Grounding}

Observed entities are grounded into analytic concepts and used to update object and scene
memory. For each detected object, the agent localizes semantic parts, matches them to
templates in $\mathcal{M}^{temp}$, and instantiates a grounded analytic concept
$\hat{O}_k$ with part parameters, poses, affordances, and candidate manipulation
templates. Details of the tool-augmented perception and grounding pipeline are provided in
Appendix~\ref{app:perception_grounding}.

A key step is object identity resolution. Given a newly grounded object $\hat{O}_k$, the
agent matches it to existing object memory entries through a hierarchical filtering
process. It first applies category and part-level consistency checks:
\[
\mathcal{C}_k = \mathrm{Filter}_{cat,part}(\hat{O}_k,\mathcal{M}^{obj}).
\]
Entries that fail these checks are discarded. For each remaining candidate
$m^{obj}_j \in \mathcal{C}_k$, the agent computes
\[
S_{k,j} =
\lambda_g S_{\mathrm{geom}} +
\lambda_T S_{\mathrm{pose}} +
\lambda_r S_{\mathrm{rel}} .
\]
If the best score exceeds a threshold, the observation is merged with the corresponding
persistent identity; otherwise, a new object entry is registered.

After identity resolution, the agent updates the scene graph with persistent object nodes
and observed spatial, visibility, containment, support, and state relations.

\subsubsection{Episode-Level Transition and Skill}

After executing a subtask or episode, the agent consolidates observed outcomes into
transition and skill memories. For each executed skill or primitive action applied to a
grounded object part, the agent records a transition entry
\[
m^{trans}=(id_k, P_{k,i}, u, s_{\mathrm{pre}}, s_{\mathrm{post}}, r),
\]
where $id_k$ is the persistent target object, $P_{k,i}$ is the grounded target part, $u$
is the executed skill or primitive action, $s_{\mathrm{pre}}$ and $s_{\mathrm{post}}$
are the states before and after execution, and $r$ records the execution result. These
entries link action effects directly to object-memory AC instances and provide
state-change knowledge for later precondition checking and effect prediction.

Successful template-based executions are further abstracted into template-grounded
skills. Specifically, when a manipulation function $f \in \mathcal{F}_{\tau_{k,i}}$
is successfully applied to a grounded part $P_{k,i}$, the agent stores a skill entry with
the skill identifier $u$, associated template type $\tau_{k,i}$, target schema, bound
parameters $\Phi$, execution backend $\pi_{\mathrm{exec}}$, outcome statistics $\eta$,
and links to the corresponding transition entries. Learned policies, such as finetuned
VLA models, are stored as policy-grounded skills together with their input prompts,
applicability conditions, execution parameters, and outcome statistics.

Through this update process, scene-specific AC instances are converted into reusable
transition and skill knowledge. The resulting memory supports state-consistent retrieval,
object re-identification, and skill reuse in future long-horizon tasks.

\subsection{Structured Memory Retrieval}

Memory entries are indexed by object-centric and manipulation-aware keys, including
object identities, part types, affordances, geometric signatures, scene relations,
object-part-action tuples, applicability conditions, and execution statistics. Given
$(L,o_t,l_i)$, retrieval follows a coarse-to-fine process:
\[
\left\{
\begin{aligned}
\mathcal{C}_t&=\mathrm{Retrieve}_{coarse}(L,o_t,l_i),\\
\mathcal{R}_t&=\mathrm{Retrieve}_{fine}(\mathcal{C}_t,\mathcal{M}_t).
\end{aligned}
\right.
\]
The coarse stage identifies candidate categories, templates, and manipulation contexts,
while the fine stage retrieves object, scene, transition, and skill entries. Fine retrieval
uses hard filters such as category, part, and affordance consistency, followed by ranking
based on geometry, pose, scene state, transition applicability, and skill statistics.

\subsection{Memory-Grounded Reasoning and Execution}

After retrieval, the agent integrates the retrieved object, scene, transition, and skill
entries with the current subtask to select an executable action plan. The reasoning
process checks whether candidate skills are applicable under the current scene state and
whether their predicted effects are consistent with the task goal.

For a candidate skill $u$, the agent first checks its precondition against the current
scene state:
\[
\mathrm{Applicable}(u, G_t)=\mathbb{I}[pre(u)\subseteq state(G_t)].
\]
If the skill is applicable, transition memory is used to estimate its expected effect:
\[
\hat{s}_{t+1}=\mathrm{Apply}(state(G_t), effect(u)).
\]
This allows the agent to avoid redundant or inconsistent actions. For example, if the
scene memory already records a microwave door as \texttt{open}, the agent should not
select another \texttt{open-door} skill before placing the object inside.

Once a grounded subtask is determined, the agent selects an executable skill from
$R^{skill}_t$. Template-grounded skills are instantiated from manipulation templates with
object-specific poses, part parameters, and execution constraints, and are executed by a
motion planner or controller. Policy-grounded skills invoke learned policies, such as
finetuned VLA models, for complex or perception-driven behaviors that are difficult to
represent with explicit templates.

The selected skill produces a low-level action sequence
\[
\boldsymbol{a}_i = \{a_{i,1},\ldots,a_{i,K_i}\},
\]
which is executed in the environment. The observed outcome $o_{t+1}$ is then compared
with the predicted postcondition. If the outcome is verified, object states, scene
relations, transitions, and skill statistics are updated accordingly. If execution fails
or the observed state contradicts the prediction, the corresponding skill attempt is
recorded as unsuccessful and the agent can re-query memory for an alternative skill. This
closes the loop between structured retrieval, state-consistent reasoning, physical
execution, and memory update. The overall closed-loop procedure is summarized in Algorithm~\ref{alg:closed_loop_acm}.

\section{Experiments}

% We evaluate our method from four perspectives: memory-dependent embodied manipulation, concept-level generalization, real-world memory capability, and component-wise ablations. 

\subsection{Memory-dependent Manipulation Results}
\label{sec:exp:main}

This section answers: \emph{Does memory improve agents on memory-dependent manipulation tasks?}

\paragraph{Benchmark.}
We evaluate on \textit{RMBench}~\citep{chen2026rmbench}, a simulation benchmark comprising nine manipulation tasks with varying memory demands, designed to support controlled evaluation across different levels of task memory complexity.

\paragraph{Baselines.}
We compare against representative baselines aligned with the table columns. \textbf{DP} denotes non-pretrained baselines. $\boldsymbol{\pi_{0.5}}$~\citep{black2025pi_} is a reactive finetuned VLA that acts only on the current observation. \textbf{Mem-0} and \textbf{Mem-VLA}~\citep{memoryvla} are memory-oriented robotic VLA policies. \textbf{MemER}~\citep{sridhar2025memer} follows a dual-system design with keyframe retrieval. For our variants, \textbf{Ours(CS)} uses only Concept-Grounded Skills, while \textbf{Ours(DP)} uses a pre-trained DP as the skill executor (distinct from the \textbf{DP} baseline column).

\paragraph{Results.}
As shown in Table~\ref{tab:rmbench_sr}, our method achieves the best overall performance
on RMBench. Compared
with reactive policies such as DP and $\pi_{0.5}$, our method benefits from persistent
object, scene, and task-progress memory rather than relying only on the current
observation. Compared with memory-augmented VLA baselines and MemER-style experience
retrieval, our analytic concept-centric memory explicitly indexes structural representations enabling the agent to retrieve state- and action-relevant entries instead of only visually similar contexts.
This leads to more consistent subtask selection and fewer redundant or physically
inconsistent actions, especially on memory-demanding tasks. The complementary performance of Ours(CS)
and Ours(DP) further suggests that our memory can support both concept-grounded skills
and policy-based execution backends.

\begin{table}[t]
\centering
\scriptsize
\setlength{\tabcolsep}{3pt}
\renewcommand{\arraystretch}{1.12}
\resizebox{\columnwidth}{!}{%
\begin{tabular}{l @{}|@{} c c c c c c c}
\toprule
\textbf{Task} & DP & $\pi_{0.5}$ & {Mem-0}$^*$ & {Mem-VLA}$^*$ & MemER & \cellcolor{blue!20}{Ours(CS)} & \cellcolor{blue!20}{Ours(DP)}  \\
\midrule
% Obs+Pick & 1\% & {9}\% & 4\% & 9\% & -- & 15\% & -- \\
Rearr. Blks & 0\% & 13\% & 89\% & 20\% & 24\% & \textbf{95\%} & \underline{90\%} \\
Put Back & 0\% & 11\% & \underline{90\%} & 35\% & 42\% & 86\% & \textbf{92\%} \\
Swap Blks & 11\% & 24\% & 67\% & 42\% & 51\% & \textbf{74\%} & \underline{72\%} \\
Swap T & {20}\% & 15\% & 14\% & 55\% & 60\% & \underline{73\%} &  \textbf{85\%}\\
Battery Try & 10\% & 16\% & 28\% & 12\% & 14\% & \underline{45\%} & \textbf{51\%} \\
Rank Try & 10\% & 6\% & 18\% & 0\% & 2\% & \textbf{26\%} &  \underline{29\%}\\
Cover Blks & 0\% & 0\% & \underline{68\%} & 31\% & 25\% & 64\% & \textbf{69\%} \\
% Press Btn & 0\% & 0\% & 0\% & 0\% & -- & -- & -- \\
\midrule
\textit{Total Avg} & 7\% & 12\% & 53\%  & 28\% & 31\% & \underline{66\%} & \textbf{70\%} \\
\bottomrule
\end{tabular}%
}
\caption{Success rate (SR, \%) on RMBench. \textit{Total Avg} is the average across all tasks. $^*$ denotes manually implemented testing.}
\label{tab:rmbench_sr}
\end{table}

\subsection{Concept Generalization}
\label{sec:exp:generalization}

This section answers: \emph{Does analytic concept-centric memory improve manipulation knowledge transfer across articulated objects that share similar part-level structures and affordances?}

\paragraph{Benchmark and task families.}
We evaluate on articulated object manipulation tasks from \textit{PartNet-Mobility}~\citep{xiang2020sapien}, following a commonly used generalization protocol (e.g., Where2Act~\citep{mo2021where2act}). The focus is concept-level transfer across articulated objects that share similar structure and affordances.
The experimental details are provided in Appendix~\ref{app:concept_generalization_details}.

\paragraph{Task Settings.}
We evaluate two transfer settings: \textbf{cross-instance transfer}, which tests generalization to unseen instances within the same category (with varying geometry/scale/pose), and \textbf{cross-object transfer}, which tests transferring part-level manipulation knowledge across categories with similar articulation/affordances. 
% In Table~\ref{tab:concept_gen_sr} (Table~2), \textit{Microwave}, \textit{Storage Furniture}, and \textit{Refrigerator} correspond to cross-instance transfer, while \textit{Washing Machine}, \textit{Kitchen Pot}, and \textit{Safe} correspond to cross-object transfer.

\paragraph{Baselines.}
We compare against representative MLLM-based manipulation baselines, \textbf{ManipLLM}~\citep{li2024manipllm} and \textbf{A3VLM}~\citep{huang2024a3vlm}, which use multimodal reasoning to infer affordances and actions from RGB-D observations and language inputs. We also include a \textbf{Semantic-memory} baseline that stores object categories and affordances but lacks explicit part-structured templates and manipulation knowledge. In contrast, our method augments MLLM reasoning with physically grounded analytic concepts and structured memory for retrieval and reuse.

\paragraph{Results.}
Table~\ref{tab:concept_gen_sr} summarizes concept generalization results under both transfer settings. We report success rate (SR, \%) on unseen instances/categories, averaged within each category.
Overall, our approach consistently improves SR across all categories. For \textit{cross-instance transfer}, we outperform the strongest baseline (typically A3VLM), indicating better robustness to geometric and pose variations within a category.
\textit{Cross-object transfer} is more challenging for prior methods (e.g., A3VLM reaches at most 40.0 SR), yet our gains are larger: +8.1 on \textit{Washing Machine}, +3.1 on \textit{Kitchen Pot}, and +10.6 on \textit{Safe}. Notably, \textbf{Semantic-memory} lags far behind, suggesting that storing coarse category/affordance cues alone is insufficient; explicitly modeling reusable part-level templates and action effects is key to transferring manipulation knowledge across categories.

\begin{table}[t]
\centering
\scriptsize
\setlength{\tabcolsep}{5pt}
\renewcommand{\arraystretch}{1.12}
\begin{tabular}{l c c c c}
\toprule
\textbf{Category} & ManipLLM & A3VLM & Semantic-mem & Ours \\
\midrule
\multicolumn{5}{l}{\textit{Cross-instance transfer}} \\
\midrule
Microwave & 41.5 & 49.7 & 20.4 & \textbf{52.2} \\
Storage Furniture & 47.0 & 53.3 & 10.5 & \textbf{60.3} \\
Refrigerator & 36.4 & 40.8 & 18.5 & \textbf{45.9} \\
\midrule
\multicolumn{5}{l}{\textit{Cross-object transfer}} \\
\midrule
Washing Machine & 19.1 & 22.3 & 8.6 & \textbf{30.4}\\
Kitchen Pot & 12.4 & 11.0 & 2.5 & \textbf{15.5}\\
Safe & 37.0 & 40.0 & 10.5 & \textbf{50.6} \\
\bottomrule
\end{tabular}
\caption{Concept generalization results. We report success rate (SR, \%) on unseen instances/categories.}
\label{tab:concept_gen_sr}
\end{table}

\subsection{Real-World Memory Evaluation}
\label{sec:exp:real}

% This section evaluates \emph{memory capability in realistic environments}, rather than only raw robot performance.

We design five real-world tabletop tasks to evaluate memory capability beyond simulation. Detailed protocols and task descriptions are provided in
Appendix~\ref{app:real_robot_setup}. These tasks require the agent to maintain object
identity, track scene relations and object locations, remember searched regions, reason
over state transitions, and retrieve reusable skills. We compare our method with a
MemER-style keyframe retrieval baseline, which stores and retrieves visual keyframes but
does not explicitly structure memory by object concepts, scene states, transitions, or
skills. We report task success rate, retrieval accuracy, and retrieval efficiency.

\paragraph{Results.}
As shown in Table~\ref{tab:real_world_results}, our method improves the average success rate from 56\% to 84\% over the keyframe retrieval baseline, while also achieving higher
retrieval accuracy and lower retrieval effort. The improvement is most evident in tasks
that require structured scene and object memory. This is because
keyframe retrieval mainly retrieves visually similar observations and reasoning, whereas our memory
explicitly indexes persistent object identities, scene relations, state transitions, and executable skills. As a result, the agent can retrieve task-relevant memory entries more
precisely, verify state-dependent preconditions, and select appropriate skills with fewer candidate inspections. Moreover, the structured representation narrows the retrieval
search space, which also improves retrieval efficiency in real-world execution.

\begin{table}[t]
\centering
\scriptsize
\setlength{\tabcolsep}{1.5pt}
\renewcommand{\arraystretch}{1.0}
\begin{tabularx}{\columnwidth}{@{} p{0.25\columnwidth} *{6}{>{\centering\arraybackslash}X} @{} }
\toprule
\multirow{2}{*}{\textbf{Task}} 
& \multicolumn{3}{c}{\textbf{Keyframe Retrieval }} 
& \multicolumn{3}{c}{\textbf{Ours}} \\
\cmidrule(lr){2-4} \cmidrule(lr){5-7}
& \textbf{SR $\uparrow$} & \textbf{RA $\uparrow$} & \textbf{RE $\downarrow$}
& \textbf{SR $\uparrow$} & \textbf{RA $\uparrow$} & \textbf{RE $\downarrow$} \\
\midrule
Slice Bread
& 75\% & -- & 3.4 
& 80\% & -- & 0.9 \\
Search
& 40\% & -- & 8.9 
& 75\% & -- & 1.4 \\
Return
& 80\% & 100\% & 3.0 
& 90\% & 100\% & 1.3 \\
Cube to Cup
& 30\% & 30\% & 3.5 
& 90\% & 100\% & 1.5 \\
Replace Block
& 55\% & 75\% & 3.8 
& 85\% & 95\% & 1.4 \\
\midrule
\textbf{Avg.}
& 56\% & 68\% & 4.5 
& \textbf{84\%} & \textbf{98\%} & \textbf{1.3} \\
\bottomrule
\end{tabularx}
\caption{
Real-world tabletop manipulation results. We compare our analytic concept-centric memory
with a MemER-style keyframe retrieval baseline. 
}
\label{tab:real_world_results}
\end{table}

\subsection{Ablation Study}
\label{sec:exp:ablation}

This section answers: \emph{How sensitive is our method to the choice of LLM model used for high-level reasoning?}

\paragraph{LLM model sensitivity.}
We vary the LLM model used by the high-level reasoning modules (e.g., subtask decomposition, concept/skill selection, and memory query formulation), while keeping all non-LLM components fixed.  The ablation settings are provided in Appendix~\ref{sec:ablation_api_models}

\begin{table}[t]
\centering
\scriptsize
\setlength{\tabcolsep}{4pt}
\renewcommand{\arraystretch}{1.12}
\begin{tabular}{l c c c}
\toprule
\textbf{LLM model} & \textbf{RMBench SR} $\uparrow$ & \textbf{Retrieval acc.} $\uparrow$ & \textbf{Obj. re-id acc.} $\uparrow$ \\
\midrule
GPT-5.5 (main) & 69\% & 95\% & 80\% \\
Gemini-3.1-pro & 65\% & 89\% & 75\% \\
GPT-4o & 67\% & 91\% & 80\% \\
Qwen-3.5 & 61\% & 88\% & 75\% \\
\bottomrule
\end{tabular}
\caption{Sensitivity to the LLM model used in high-level reasoning. We report success rate on RMBench, retrieval accuracy, and object re-identification accuracy.}
\label{tab:ablation_api_models}
\end{table}
Table~\ref{tab:ablation_api_models} shows that the choice of reasoning LLM measurably affects both task performance (RMBench SR) and memory reliability (retrieval / re-identification). Stronger models improve SR and retrieval accuracy, suggesting that better subtask decomposition and more precise memory queries translate into more correct object/skill retrieval and, in turn, more state-consistent execution. We also observe that object re-identification is less variable than retrieval accuracy across models, indicating that identity resolution benefits primarily from the structured analytic concept representation and matching signals, while the LLM mainly influences which entities and skills are queried and selected..

\section{Conclusion}

We introduced an analytic concept-centric memory framework for long-horizon agentic embodied manipulation. By organizing memory around reusable analytic templates, grounded object instances, scene states, action transitions, and executable skills, the framework supports structured retrieval, state-consistent reasoning, object re-identification, and skill reuse.

Experiments on memory-dependent manipulation, concept generalization, real-world tasks, and ablations show that our method improves task completion, retrieval correctness, and cross-object skill reuse over unstructured or embedding-based memory alternatives. Future work will scale the template library to deformable objects, improve robustness to perception noise and state drift, and explore learning-based retrieval while preserving the interpretability of analytic concepts.

% This document does not cover the content requirements for ACL or any
% other specific venue.  Check the author instructions for
% information on
% maximum page lengths, the required ``Limitations'' section,
% and so on.

% \section*{Acknowledgments}

% This document has been adapted
% by Steven Bethard, Ryan Cotterell and Rui Yan
% from the instructions for earlier ACL and NAACL proceedings, including those for
% ACL 2019 by Douwe Kiela and Ivan Vuli\'{c},
% NAACL 2019 by Stephanie Lukin and Alla Roskovskaya,
% ACL 2018 by Shay Cohen, Kevin Gimpel, and Wei Lu,
% NAACL 2018 by Margaret Mitchell and Stephanie Lukin,
% Bib\TeX{} suggestions for (NA)ACL 2017/2018 from Jason Eisner,
% ACL 2017 by Dan Gildea and Min-Yen Kan,
% NAACL 2017 by Margaret Mitchell,
% ACL 2012 by Maggie Li and Michael White,
% ACL 2010 by Jing-Shin Chang and Philipp Koehn,
% ACL 2008 by Johanna D. Moore, Simone Teufel, James Allan, and Sadaoki Furui,
% ACL 2005 by Hwee Tou Ng and Kemal Oflazer,
% ACL 2002 by Eugene Charniak and Dekang Lin,
% and earlier ACL and EACL formats written by several people, including
% John Chen, Henry S. Thompson and Donald Walker.
% Additional elements were taken from the formatting instructions of the \emph{International Joint Conference on Artificial Intelligence} and the \emph{Conference on Computer Vision and Pattern Recognition}.


\bibliography{custom}

\clearpage
\appendix
\section{Implementation Details}
\subsection{Perception and Concept Grounding}
\label{app:perception_grounding}

\begin{figure*}[t]
    \centering
    \includegraphics[width=0.98\linewidth]{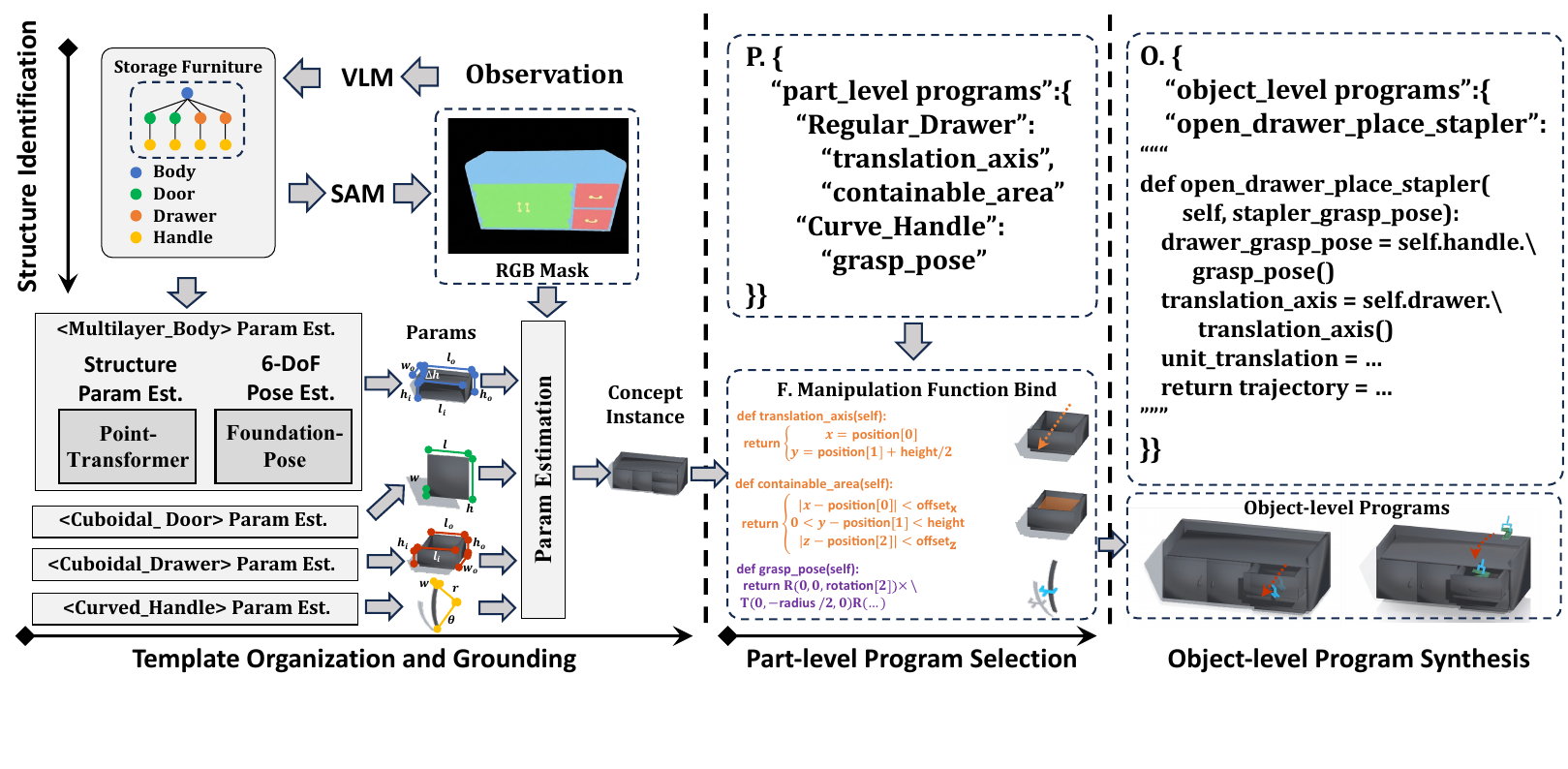}
    \caption{
    \textbf{Perception and analytic concept grounding.}
    The agent invokes visual foundation models and analytic estimators as tools to
    convert observations into executable analytic concepts. VLM-guided structure
    identification and SAM-based segmentation provide object and part masks, while
    pose and parameter estimation tools instantiate matched templates into grounded
    part concepts. The resulting part-level programs are composed into object-level
    programs for downstream manipulation.
    }
    \label{fig:concept_grounding}
\end{figure*}

We implement perception and analytic concept grounding as a tool-augmented process,
as illustrated in Fig.~\ref{fig:concept_grounding}. Instead of relying on a monolithic
perception model, the agent invokes a set of vision foundation models and analytic
estimators as external tools. Given an RGB-D observation, the agent first identifies
task-relevant objects and candidate parts with VLM-guided object/part recognition and
SAM-based~\citep{ravi2025sam} segmentation. The resulting masks are then passed to pose and parameter
estimation tools, such as FoundationPose~\citep{wen2024foundationpose} for 6D pose estimation and template-specific
parameter estimators for structural fitting.

For structural parameter estimation, we adopt a Point-Transformer~\citep{zhao2021point} encoder that extracts
$128$ groups of points with group size $32$ from the input point cloud and applies $12$
attention layers with $6$ heads each. We then apply average pooling to obtain a global
point-cloud feature, followed by a three-layer MLP with ReLU activations to regress the
template-specific structural parameters.

The grounding process is organized around template types. For each detected part, the
agent selects a candidate template type $\tau_i$ according to the semantic synopsis in
$\mathcal{M}^{sem}$ and the structural template in $\mathcal{M}^{temp}$. The perception
tools estimate scene-specific parameters and poses, denoted as
$(\hat{\Theta}_i,\hat{T}_i)$, which are then bound to the matched template
$m^{temp}_{\tau_i}$. This produces a grounded part-level analytic concept
\[
P_i=\mathcal{P}(\tau_i,\hat{\Theta}_i,\hat{T}_i,A_i,\mathcal{F}_{\tau_i}),
\]
where $A_i$ denotes the inferred part affordance and $\mathcal{F}_{\tau_i}$ denotes the
manipulation functions supported by the corresponding template type.

After part-level grounding, the agent assembles grounded parts into an object-level
analytic concept. The object-level concept stores the persistent object identity,
template-instantiated parts, estimated geometry, grounded poses, and supported
manipulation functions. These grounded concepts are written into object memory and linked
to scene memory through spatial and state relations. In this way, the output of visual
tools is not stored as raw masks or embeddings, but converted into executable analytic
concept entries that can be retrieved, reasoned over, and reused by the embodied agent.

In addition to structural grounding, the same template linkage is used to derive
part-level and object-level programs. Part-level programs expose reusable functions
defined by the matched templates, such as estimating a drawer translation axis,
computing a containable region, or generating a handle grasp pose. Object-level programs
compose these part-level functions into executable routines, such as opening a drawer and
placing an object inside. Thus, concept grounding provides both a structured memory entry
and a programmatic interface for downstream skill selection and execution.

Our method does not assume perfect perception. Missing, low-confidence, or inconsistent
tool outputs are treated as partial observations and do not overwrite existing object
memory unless verified by later observations or execution feedback. This design allows
the agent to benefit from strong vision foundation models while keeping memory explicit,
structured, and updateable.

\subsection{Concept Generalization: Experimental Details}
\label{app:concept_generalization_details}

This subsection provides additional details for the concept generalization experiments
in Section~\ref{sec:exp:generalization}.

\paragraph{Environments and categories.}
We evaluate articulated object manipulation in simulation using the SAPIEN
simulator~\citep{xiang2020sapien}, with tasks drawn from \textit{ManiSkill} and assets
from \textit{PartNet-Mobility}. Each episode contains a single articulated target
object with a manipulable part (e.g., a door, lid, or handle) whose kinematic state is
observable only through perception and interaction.

% \paragraph{Transfer protocols.}
% We consider two transfer settings.
% (i) \textbf{Cross-instance transfer} holds the object category fixed and evaluates on
% unseen object instances within that category, with variation in geometry, scale, and
% part pose.
% (ii) \textbf{Cross-object transfer} evaluates on a different category that shares a
% compatible part-level structure and affordance (e.g., hinged doors or pullable lids),
% testing whether retrieved part templates, transitions, and skills remain applicable.
% In both settings, all library entries used at test time follow the same schemas as in
% the main system; only the object instances and/or categories differ.

\paragraph{Episode initialization.}
In each trial, the target object is placed at the center of the scene. The articulated
joint configuration is randomized, with a 50\% probability of starting from the fully
closed state and a 50\% probability of starting from an open configuration sampled
randomly along the joint's range of motion. We use an RGB-D camera with known intrinsic
parameters that points toward the scene center and is positioned on the upper
hemisphere, with azimuth sampled uniformly from $[0^{\circ}, 360^{\circ})$ and elevation
sampled uniformly from $[30^{\circ}, 60^{\circ}]$.

\paragraph{Success criteria and metrics.}
We evaluate with success rate (SR, \%), defined as the ratio of successful trials to the
total number of test trials. A trial is considered successful if the target joint's
movement exceeds 0.01 (in simulator length units) or exceeds 0.5 when normalized by the
joint's maximum motion range. SR is averaged over evaluation episodes and then averaged
within each category as reported in Table~\ref{tab:concept_gen_sr}. We use identical
success predicates for all methods to ensure comparability.

\paragraph{Interaction budget.}
We adopt an interaction budget of 5 for each action proposal.

\paragraph{Reporting.}
Unless otherwise stated, we use the same prompting templates and output schemas across
methods within each setting, and we aggregate results across randomized initializations
to reduce sensitivity to a single pose or viewpoint.

\subsection{Real-Robot Experimental Setup}
\label{app:real_robot_setup}

\begin{figure*}[t]
    \centering
    \includegraphics[width=0.98\linewidth]{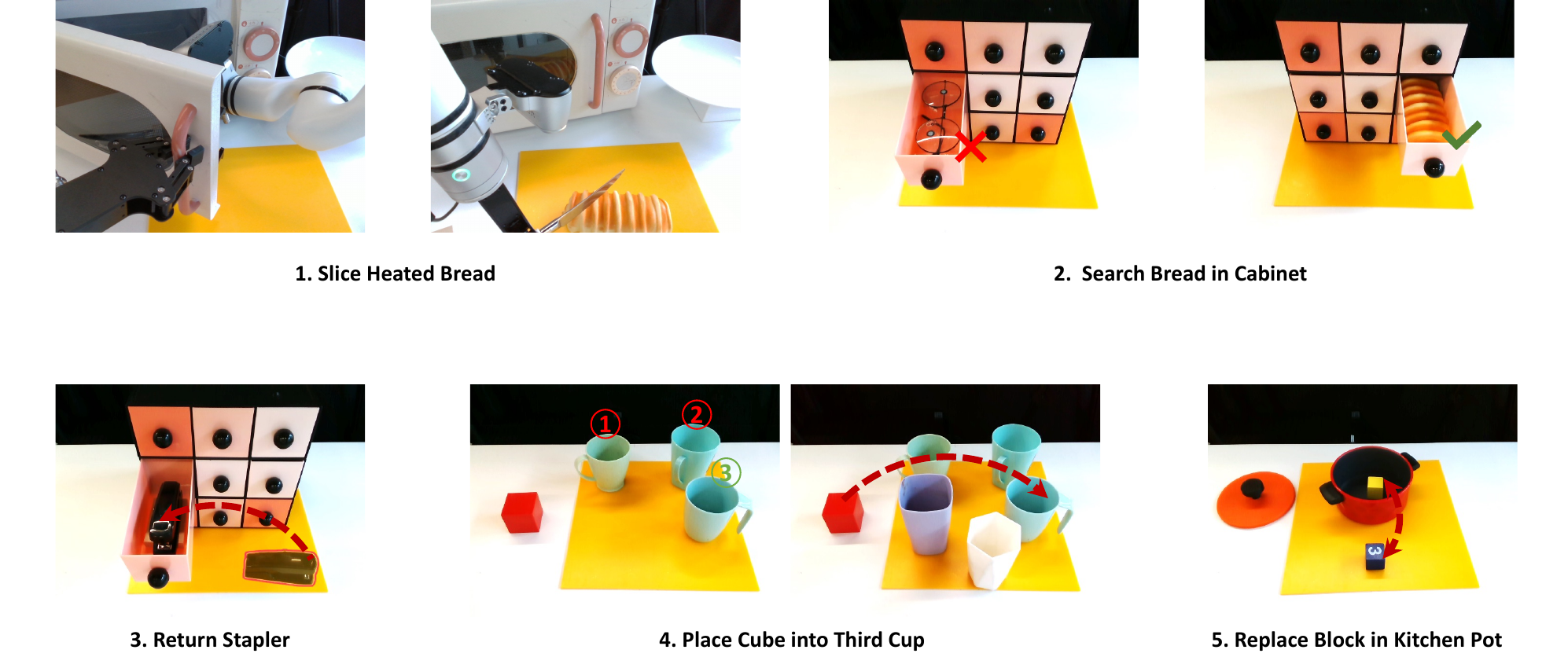}
    \caption{Real-world experimental setup for the tabletop manipulation tasks in Section~\ref{sec:exp:real}.}
    \label{fig:real_robot_setup}
\end{figure*}

\paragraph{Perception.}
We use five RGB-D cameras (Intel RealSense D435): two wrist-mounted cameras, one
egocentric (end-effector) camera, and two third-person cameras.

\paragraph{Compute.}
The system runs on a workstation equipped with an Intel Core i9-14900K processor, 64GB
of RAM, and an NVIDIA RTX 4090 GPU, enabling real-time inference and planning.

\paragraph{Metrics.}
In addition to success rate, we emphasize retrieval correctness and efficiency:
\begin{itemize}
    \item \textbf{Success rate (SR):} fraction of episodes with full task completion within the episode horizon.
    \item \textbf{Retrieval accuracy:} for each subtask-level retrieval query, we mark it correct only if \emph{both} (i) the retrieved target object identity matches the intended object instance and (ii) the retrieved skill (template- or policy-grounded) matches the intended manipulation procedure. Retrieval accuracy is the mean correctness over all subtask queries.
    \item \textbf{Retrieval efficiency:} (i) average end-to-end query latency.
\end{itemize}

\paragraph{Tasks.}
We evaluate on five real-world tabletop tasks (Table~\ref{tab:real_world_tasks}), each specified by a preceding context task and a target task that stresses particular memory capabilities.

\begin{table*}[t]
\centering
\small
\begin{tabular}{p{0.18\textwidth} p{0.78\textwidth}}
\toprule
\textbf{Task} & \textbf{Slice Heated Bread} \\
\textbf{Preceding task} & Heat the bread in a microwave. \\
\textbf{Task description} & Retrieve the heated bread, place it on the cutting board, and slice it with the knife. \\
\textbf{Main challenge} 
&  Scene memory for bread location; skill memory for microwave operation and slicing. \\
\midrule
\textbf{Task} & \textbf{Search Bread in Cabinet} \\
\textbf{Preceding task} & -- \\
\textbf{Task description} & Search the cabinet and find the bread. \\
\textbf{Main challenge} & Scene memory for searched locations. \\
\midrule
\textbf{Task} & \textbf{Return Stapler} \\
\textbf{Preceding task} & Observe the stapler at its initial location and move it to another workspace region. \\
\textbf{Task description} & Put the stapler back to its original location. \\
\textbf{Main challenge} & Object memory for persistent identity; scene memory for original location. \\
\midrule
\textbf{Task} & \textbf{Place Cube into Third Cup} \\
\textbf{Preceding task} & Arrange multiple cups in order; optionally change cup positions. \\
\textbf{Task description} & Pick up the cube and place it into the third cup. \\
\textbf{Main challenge} & Transition memory for tracking cup order; object memory for target cup disambiguation. \\
\midrule
\textbf{Task} & \textbf{Replace Block in Kitchen Pot} \\
\textbf{Preceding task} & A block with a random color is placed into the Kitchen Pot. \\
\textbf{Task description} & Replace the block in the Kitchen Pot with another specified color block.  \\
\textbf{Main challenge} & Scene memory for the hidden block location. \\
\bottomrule
\end{tabular}
\caption{Real-world tabletop tasks for evaluating analytic concept-centric memory. The tasks cover different memory capabilities, including object identity, scene relations, state transitions, search history, and reusable skill selection.}
\label{tab:real_world_tasks}
\end{table*}

\subsection{Ablation protocol (Table~\ref{tab:ablation_api_models}).}\label{sec:ablation_api_models}
We change only the LLM used for high-level reasoning (subtask decomposition, concept/skill selection, and memory query formulation). All non-LLM components (perception/grounding, structured memory indexing, retrieval, identity resolution, and low-level execution) are kept fixed. We use the same prompts and output schemas across models to ensure a controlled comparison.

% \paragraph{Metric computation.}
% \textbf{RMBench SR} is computed as the mean success rate over RMBench tasks (as in Table~\ref{tab:rmbench_sr}). \textbf{Retrieval accuracy} and \textbf{object re-identification accuracy} follow the definitions in Section~\ref{sec:exp:real}: retrieval is evaluated at the subtask level (object + skill jointly correct), and re-identification is evaluated on revisit events that require persistent identity assignment.

\section{Atomic Manipulation Primitives}
\label{sec:appendix_primitives}
The manipulation functions in $\mathcal{F}_{\tau}$ are implemented by composing a
small set of \,\emph{atomic} manipulation primitives (Table~\ref{tab:atomic_primitives}).
Each primitive corresponds to a low-level, robot-executable operation (e.g., reaching,
grasping, or contact-driven pushing) with a standardized interface over grounded
objects/parts and geometric parameters. Given a grounded analytic concept
$P=\mathcal{P}(\tau,\Theta,T,A,\mathcal{F}_{\tau})$, a template function
$f\in\mathcal{F}_{\tau}$ first derives the required arguments (target poses, axes,
or goal positions) from $(\Theta,T,A)$ and the current scene state, and then invokes the
appropriate primitive(s). More complex behaviors---such as opening an articulated
component or inserting an object---are realized as short sequences of primitives with
intermediate state checks, enabling consistent execution and outcome logging for
transition and skill memory updates.

% \section{Potential Risks}

% This work focuses on controlled simulation and tabletop manipulation. Potential risks
% mainly arise from the broader deployment of memory-enabled robotic agents. In less
% controlled environments, incorrect retrieval or inappropriate skill execution may cause
% unintended object movement, object damage, or unsafe robot behavior. Such systems should
% therefore be used with workspace constraints, execution monitoring, and human oversight.

% Memory-enabled agents may also store information about objects, scene layouts, and user
% environments. Although our experiments only record task-relevant object states and
% manipulation outcomes, real-world deployments should include memory management policies,
% such as limiting stored content, supporting memory deletion, and protecting stored
% information from unauthorized access. Our system is intended as a research framework for
% controlled embodied manipulation, not for safety-critical or human-facing deployment.

\begin{algorithm*}[t]
\small
\caption{Closed-Loop Analytic Concept-Centric Memory Agent}
\label{alg:closed_loop_acm}
\begin{algorithmic}[1]
\Require Stable semantic library $\mathcal{M}^{sem}$, template library $\mathcal{M}^{temp}$
\State Initialize dynamic memories 
$\mathcal{M}^{obj}, \mathcal{M}^{scene}, \mathcal{M}^{trans}, \mathcal{M}^{skill}$
\State $\mathcal{M} \leftarrow \{\mathcal{M}^{sem},\mathcal{M}^{temp},
\mathcal{M}^{obj},\mathcal{M}^{scene},\mathcal{M}^{trans},\mathcal{M}^{skill}\}$

\While{agent is active}
    \State Receive an episode instruction $L$
    \State Observe current scene $o_t$

    \Statex \Comment{\textbf{Memory-guided task parsing}}
    \State Retrieve relevant past objects, states, transitions, and skills from $\mathcal{M}$
    \State Decompose $L$ into subtasks $\mathbf{l}=\{l_1,\ldots,l_N\}$ conditioned on retrieved memory

    \For{each subtask $l_i \in \mathbf{l}$}
        \State Observe current scene $o_t$

        \Statex \Comment{\textbf{Phase I: Memory construction and grounding}}
        \State Parse $o_t$ to obtain candidate objects, parts, and scene relations
        \State Invoke perception tools for detection, segmentation, pose estimation, and parameter estimation
        \For{each detected part}
            \State Match it to a template type $\tau$ using $\mathcal{M}^{sem}$ and $\mathcal{M}^{temp}$
            \State Instantiate grounded part concept
            \[
            P_{k,j}=\mathcal{P}(\tau_{k,j},\hat{\Theta}_{k,j},\hat{T}_{k,j},
            A_{k,j},\mathcal{F}_{\tau_{k,j}})
            \]
        \EndFor
        \State Resolve persistent object identities and update object memory $\mathcal{M}^{obj}$
        \State Update scene memory $\mathcal{M}^{scene}$ with object states and scene relations

        \Statex \Comment{\textbf{Phase II: Memory-grounded execution}}
        \State Retrieve task-relevant memory entries:
        \[
        \mathcal{R}_t =
        \mathrm{Retrieve}_{fine}(
        \mathrm{Retrieve}_{coarse}(L,o_t,l_i),\mathcal{M})
        \]
        \State Obtain candidate objects, scene states, transitions, and skills from $\mathcal{R}_t$
        \For{each candidate skill $u$}
            \State Check preconditions with current scene state:
            \[
            \mathrm{App}(u,G_t)=\mathbb{I}[pre(u)\subseteq state(G_t)]
            \]
            \State Predict expected postcondition using transition memory
        \EndFor
        \State Select an applicable skill $u^\ast$ consistent with subtask $l_i$
        \State Instantiate $u^\ast$ with grounded object and part parameters
        \State Send the resulting low-level action sequence to the robot executor

        \Statex \Comment{\textbf{Feedback and memory update}}
        \State Observe execution feedback and updated observation $o_{t+1}$
        \If{the predicted postcondition is verified}
            \State Update object and scene states in $\mathcal{M}^{obj}$ and $\mathcal{M}^{scene}$
            \State Record transition
            \[
            m^{trans}=(id_k,P_{k,j},u^\ast,
            s_{\mathrm{pre}},s_{\mathrm{post}},r)
            \]
            \State Update skill statistics in $\mathcal{M}^{skill}$
        \Else
            \State Mark the skill attempt as failed and update skill statistics
            \State Re-query memory or select an alternative skill in the next loop
        \EndIf
        \State $t \leftarrow t+1$
    \EndFor
\EndWhile
\end{algorithmic}
\end{algorithm*}

\begin{table*}[t]
\centering
\small
\begin{tabular}{p{0.18\linewidth} p{0.75\linewidth}}
\toprule
\textbf{Primitive} & \textbf{Definition} \\
\midrule

\texttt{grasp(obj)} &
The robot establishes a stable grasp on the target object by closing the gripper or applying a holding constraint through the end-effector. The action results in the object becoming attached to the manipulator for subsequent manipulation. \\

\midrule

\texttt{reach(obj)} &
The manipulator moves toward the target object while maintaining the current gripper state unchanged. The action terminates before grasping, contact-driven translation, or any interaction that changes the object state occurs. \\

\midrule

\texttt{move(pos, obj)} &
The manipulator translates the end-effector or a grasped object toward a target position defined relative to a reference object or workspace region, while maintaining the current grasp configuration unchanged. The action represents spatial relocation without intentional contact interaction. \\

\midrule

\texttt{flip(obj, axis)} &
The robot rotates the target object around a specified axis, typically by approximately $180^\circ$, such that the object's vertical orientation, exposed surface, or functional side becomes inverted. Unlike general rotational adjustment, this primitive emphasizes discrete orientation reversal. \\

\midrule

\texttt{rotate(obj, axis, angle)} &
The robot changes the orientation or articulated state of the target object through rotational motion without causing vertical inversion. This primitive includes in-plane pose adjustment, 3D orientation alignment, and manipulation of rotational joints or mechanisms. \\

\midrule

\texttt{pull(obj, pos)} &
The end-effector establishes pulling contact with the target object and moves backward relative to the object, generating traction that causes the object to translate toward the target position. The action terminates once object motion ceases and contact is released. \\

\midrule

\texttt{push(obj, pos)} &
The end-effector makes contact with the target object and applies forward translational force, causing the object to move toward a target position under sustained contact. The action ends when the object stops moving and the pushing contact is terminated. \\

\midrule

\texttt{insert(obj1, obj2)} &
The robot aligns object \texttt{obj1} with an opening, slot, cavity, or receptacle associated with \texttt{obj2}, and subsequently translates \texttt{obj1} along the insertion axis until partial or full engagement is achieved. The action emphasizes constrained motion under geometric alignment. \\

\midrule

\texttt{release(obj)} &
The robot removes the grasping constraint from the target object by opening the gripper or disengaging the holding mechanism. After completion, the object is no longer mechanically constrained by the manipulator and remains free in the environment. \\

\bottomrule
\end{tabular}
\caption{Definitions of atomic manipulation primitives used for analytic concept decomposition.}
\label{tab:atomic_primitives}
\end{table*}

\end{document}